\documentclass[journal]{IEEEtai}

\usepackage[colorlinks,urlcolor=blue,linkcolor=blue,citecolor=blue]{hyperref}

\usepackage{color,array}

\usepackage{graphicx}

\usepackage{xcolor}

\usepackage{amsmath}
\usepackage{amssymb}

\usepackage{comment}

\usepackage{tabularx}

\usepackage{booktabs}

\usepackage{array}

\usepackage{multirow}

\usepackage{pifont}
\newcommand{\xmark}{\ding{55}} 

\begin{document}

\title{Watermarking Language Models through \\ Language Models} 


\author{Agnibh Dasgupta, Abdullah All Tanvir, and Xin Zhong 
\thanks{Agnibh Dasgupta, Abdullah All Tanvir, and Xin Zhong are with the Department of Computer Science, University of Nebraska Omaha, Omaha, NE, 68182 USA. E-mail: (adasgupta, atanvir, xzhong)@unomaha.edu}
}



\maketitle

\begin{abstract}
Watermarking the outputs of large language models (LLMs) is critical for provenance tracing, content regulation, and model accountability. Existing approaches often rely on access to model internals or are constrained by static rules and token-level perturbations. Moreover, the idea of steering generative behavior via prompt-based instruction control remains largely underexplored. 
We introduce a prompt-guided watermarking framework that operates entirely at the input level and requires no access to model parameters or decoding logits. The framework comprises three cooperating components: a Prompting LM that synthesizes watermarking instructions from user prompts, a Marking LM that generates watermarked outputs conditioned on these instructions, and a Detecting LM trained to classify whether a response carries an embedded watermark. 
This modular design enables dynamic watermarking that adapts to individual prompts while remaining compatible with diverse LLM architectures, including both proprietary and open-weight models. We evaluate the framework over 25 combinations of Prompting and Marking LMs, such as GPT-4o, Mistral, LLaMA3, and DeepSeek. Experimental results show that watermark signals generalize across architectures and remain robust under fine-tuning, model distillation, and prompt-based adversarial attacks, demonstrating the effectiveness and robustness of the proposed approach.

\end{abstract}

\begin{IEEEImpStatement}
This work introduces a modular, prompt-guided watermarking framework designed to embed and detect imperceptible signals in large language model (LLM) outputs without requiring access to model internals. By leveraging prompt-based instruction generation, the method enables dynamic, context-aware watermarking that adapts to each user input while remaining compatible with black-box and proprietary models. The framework’s flexibility and generality make it applicable to real-world deployment scenarios where control over model weights or decoding processes is not feasible. Extensive evaluation across different Prompting–Marking LM combinations demonstrates that the watermark signal is robust to state-of-the-art attacks. This work advances the state of LLM accountability and verification, providing a practical, scalable, and model-agnostic solution for secure content provenance in generative AI systems.

\end{IEEEImpStatement}

\begin{IEEEkeywords}
Content authentication, instruction control, large language models, prompt engineering, robust watermarking
\end{IEEEkeywords}

\section{Introduction}
\label{sec:intro}
\IEEEPARstart{L}{anguage} model watermarking refers to the practice of embedding identifiable markers or signatures within the outputs of language models. These watermarks are designed to be imperceptible to end-users but detectable by an authorized system, allowing the origin of the content to be traced back to a particular model or model owner~\cite{nie2024deep}. This process typically involves embedding patterns or codes within the generated text in such a way that they do not affect its quality or meaning while enabling traceability and authentication. Recent advancements in deep learning have opened new opportunities for incorporating these watermarks efficiently within large language models (LLMs).

\begin{figure}[!htb]
    \vspace{-0.75em}
    \includegraphics[width=0.99\linewidth]{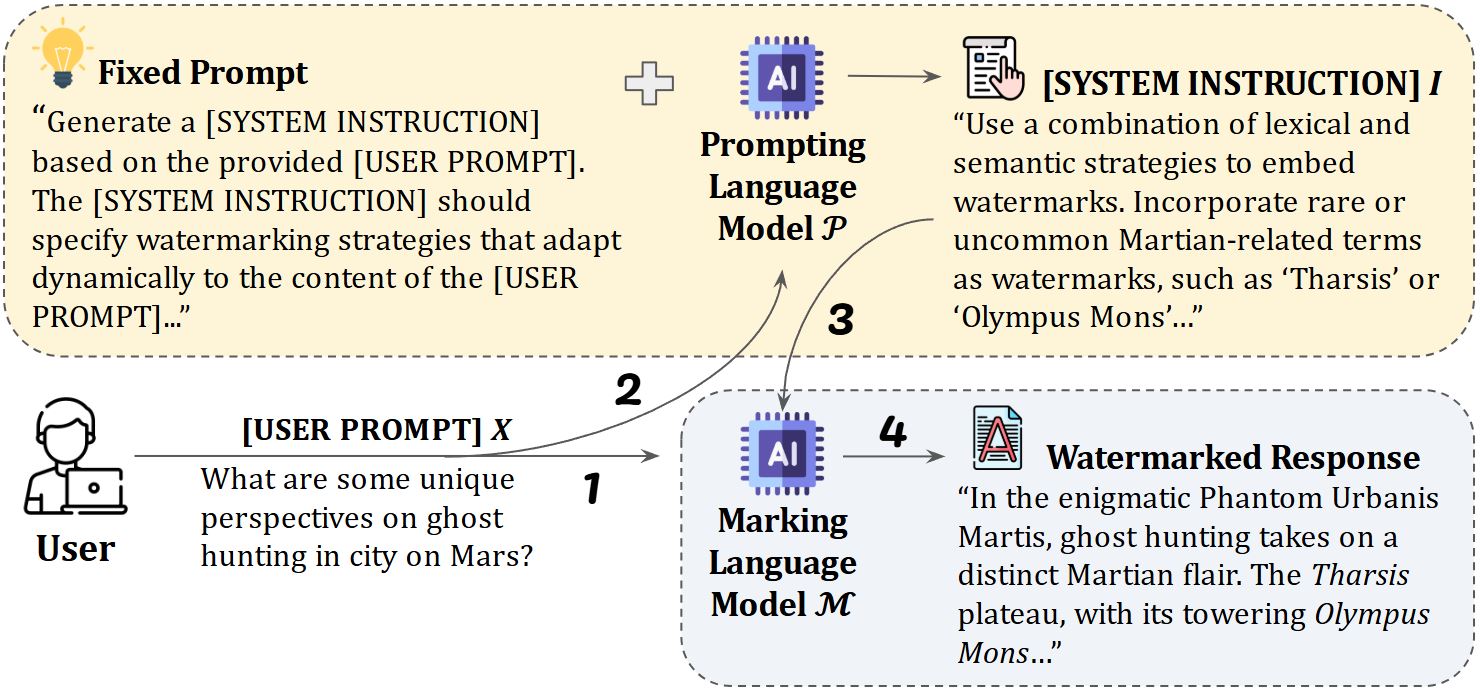}
    \vspace{-0.5em}
    \caption{Overview of our watermarking scheme. The yellow and blue boxes highlight the watermark generation and watermarking phases respectively.}
    \label{fig:Overview}
    \vspace{-0.75em}
\end{figure}

The need for language model watermarking arises due to the growing prevalence of generative AI systems in real-world applications, where issues such as content ownership, intellectual property protection, and accountability are becoming increasingly critical. As language models are deployed in diverse settings, from automated content generation to conversational agents, ensuring the integrity of the generated text is vital. Watermarking can help identify misuse of models, detect AI-generated content, and enforce responsible usage, particularly when it comes to protecting copyrighted material, preventing deepfake text, or monitoring compliance with ethical guidelines.

Recent advancements in watermarking techniques for LLMs have focused on two primary approaches: embedding watermarks directly into generated text and embedding watermarks within the LLM architecture itself. Text-based watermarking methods typically alter token probabilities or manipulate token selections during generation to make watermarks detectable by specific algorithms while maintaining imperceptibility for users. Notable contributions include signal embedding through token manipulation~\cite{kirchenbauer2023watermark, zhao2023provable}, cryptographically inspired undetectable watermarks~\cite{christ2024undetectable}, and adaptive strategies based on token entropy~\cite{liu2024adaptive}. While these methods achieve various levels of robustness, detectability, and minimal impact on text quality, model-based watermarking techniques embed watermarks within the internal structure of LLMs~\cite{li2023watermarking, liu2023watermarking}, protecting models against tampering and unauthorized fine-tuning. These methods ensure the integrity of model weights, prevent unauthorized use, and safeguard intellectual property.

Despite the progress, challenges remain in achieving dynamic, robust, and imperceptible watermarking that can adapt to different text contexts. 
Existing watermarking methods are limited by the reliance on static rules, token-level manipulations, or the need to access model internals such as logits or training weights. 
Moreover, current approaches often do not take full advantage of the instruction-following capabilities of modern LLMs to adapt watermarking behavior dynamically based on content. 
These gaps motivate the design of our prompt-guided, multi-LM watermarking framework.

To this end, we propose a novel dynamic watermarking method that leverages the internal capabilities of language models to generate and insert watermarks via prompts during content generation. 
Unlike existing methods that statically embed watermarks into text or model parameters, the proposed model has the ability to embed watermark signals adaptively per prompt, as guided by an instruction generated in context, rather than relying on a fixed watermarking rule. Thus, our approach dynamically adjusts the watermarking based on user prompts, ensuring flexibility and robustness across various applications.

To operationalize this dynamic approach, we introduce a modular framework that consists of three functionally distinct components: a Prompting LM that generates watermarking instructions from user prompts, a Marking LM that incorporates those instructions to generate watermarked responses, and a Detecting LM trained to classify whether an output carries a watermark. As shown in Figure~\ref{fig:Overview}, the process begins when a user submits a prompt to the Marking LM (Step 1). This prompt is forwarded to the Prompting LM, which also receives a fixed prompt template (Step 2). The  Prompting LM generates a prompt-specific watermarking instruction based on this template (Step 3). The Marking LM combines the original prompt with the generated instruction to produce a watermarked output (Step 4). Finally, these watermarked outputs, along with their non-watermarked variants are used to train the Detecting LM.

The proposed work makes three main contributions. 
First, we introduce a prompt-guided watermarking framework that uses a modular, multi-LM design to embed and detect imperceptible signals in generated text, requiring no access to model internals. 
Second, we demonstrate that the framework is dynamic and adaptive—generating watermarking instructions conditioned on each user prompt—while remaining compatible with diverse, black-box LLMs. 
Third, we show that the watermarking signals generalize across different model architectures and remain robust under both model-based transformations (fine-tuning and distillation) and prompt-based attacks.

\vspace{-1.0em}
\section{Related Work}
\label{sec:related}
\vspace{-0.5em}

Watermarking techniques for LLMs have gained traction as essential tools for safeguarding intellectual property and authenticating AI-generated content. Broadly, these methods can be categorized~\cite{liu2024survey} into two types: watermarking embedded in the generated text and watermarking embedded directly in the LLM's architecture.

\vspace{-0.75em}
\subsection{Watermarking in Generated Text}
\vspace{-0.25em}
Watermarking in generated text has been an effective approach to identify AI-generated content while protecting intellectual property. Several methods focus on embedding signals directly into the generated outputs of LLMs, making these watermarks detectable by algorithms but imperceptible to humans.

Kirchenbauer et al.~\cite{kirchenbauer2023watermark} introduced a watermarking framework for large language models that embeds signals into generated text, which remain invisible to humans but are algorithmically detectable from a short span of tokens. Their method leverages a randomized selection of "green" tokens to softly promote their use during sampling, thus embedding the watermark with negligible impact on text quality. The watermark detection relies on a statistical test with interpretable p-values, without requiring access to the model's API or parameters. 
Zhao et al.~\cite{zhao2023provable} proposed the Unigram-Watermark, a method for watermarking LLM-generated text by grouping tokens in a fixed manner. They provide a theoretical framework assessing watermark effectiveness, robustness, and tolerance to text editing and paraphrasing. Experiments demonstrate detection accuracy with small impact on text quality. 
Wang et al.~\cite{wang2023towards} introduced Codable Text Watermarking for LLMs, enabling watermarks to carry multi-bit customizable information, such as model version or user ID. They use a token-based Balance-Marking method, which employs a proxy language model to split the vocabulary into balanced parts. Their evaluation system includes metrics such as watermark success rate, robustness, and impact on text quality. 
Christ et al.~\cite{christ2024undetectable} proposed a cryptographically-inspired method for embedding undetectable watermarks in language model outputs by subtly altering token probabilities. Their approach requires a secret key for detection and ensures no observable change to the output distribution, maintaining the quality of the generated text. The method is based on the existence of one-way functions and is designed to remain undetectable even under adaptive querying. 
Kuditipudi et al.~\cite{kuditipudi2023robust} proposed a watermarking methodology for autoregressive language models that maps tokens to a sequence of random numbers generated with a randomized key. Watermarked text is generated by aligning the output tokens to this random number sequence. The method was applied to multiple models, demonstrating robustness to paraphrasing attacks and reliable detection from a small number of tokens, though detection performance varied across models. 
Guan et al.~\cite{guan2024codeip} introduced CodeIP, a multi-bit watermarking technique for code generated by LLMs that inserts watermarks by manipulating tokens to preserve provenance details such as vendor ID. The approach ensures syntactical correctness by constraining the token prediction process with a type predictor. Experiments across multiple programming languages demonstrated the method's effectiveness in watermarking while maintaining code correctness. 
Liu and Bu~\cite{liu2024adaptive} proposed an adaptive watermarking strategy for LLM-generated text. Their method selectively adds watermarks to high-entropy token distributions, while leaving low-entropy tokens untouched. They adaptively scale the output logits based on semantic embeddings of previously generated text, avoiding the use of fixed token lists. Experiments demonstrate robustness and security, with text quality comparable to un-watermarked models. 
In contrast to token probability manipulation, Munyer et al.~\cite{munyer2024deeptextmark} developed \textit{DeepTextMark}, which embeds watermarks by substituting lexical synonyms within the generated text. This method relies on subtle lexical changes, such as synonym replacements, that preserve the text's meaning while embedding a watermark, ensuring both robustness and imperceptibility. 
These diverse approaches illustrate a growing focus on balancing robustness, subtlety, and quality in watermarking AI-generated text.

\vspace{-0.75em}
\subsection{Watermarking within Language Model Architectures.} 
\vspace{-0.25em}
An emerging area of research focuses on embedding watermarks directly within the parameters of LLMs to safeguard the integrity of the models themselves. Unlike text-based watermarks, these techniques target the internal structure of LLMs, ensuring protection against tampering and misuse at the model level.

Li et al.~\cite{li2023watermarking} proposed a watermarking strategy that embeds watermarks in the quantization process of LLMs. Their approach allows the watermark to be active in fp32 mode while remaining hidden in int8 mode, preventing further fine-tuning of the model. The method was successfully applied to open-source models such as GPT-Neo and LLaMA, providing a potential direction for protecting model weights. 
Liu et al.~\cite{liu2023watermarking} introduced TextMarker, a backdoor-based watermarking technique that protects private information embedded in training data through membership inference. The method requires marking a small subset of the data and is effective under black-box access assumptions. Experiments show that marking as little as 0.1\% of the data enables effective membership inference with minimal impact on model performance. 
Xu et al.~\cite{xu2024instructional} introduced a lightweight fingerprinting technique for LLMs using instruction tuning. A private key is implanted as an instruction backdoor, causing the model to generate specific text when the key is present. The method is shown to be effective across 11 LLMs without affecting normal model behavior, while also providing robustness against fingerprint guessing and parameter-efficient training. 
Huang et al.~\cite{huang2024multi} introduced multi-designated detector watermarking (MDDW) for LLMs, allowing specific detectors to identify watermarked outputs while maintaining output quality for regular users. The method is based on multi-designated verifier signatures (MDVS) and supports an optional claimability feature, enabling model providers to assert ownership of outputs. Their implementation demonstrates flexibility and satisfactory performance metrics. 
Ren et al.~\cite{ren2024subtle} proposed Robust and Imperceptible Watermarking (RIW) for LLM-generated text, which leverages token prior probabilities to balance detectability and imperceptibility. The method partitions tokens into two groups based on their prior probabilities and applies different embedding strategies. A 'voted z-test' is used for watermark detection. Experimental results show that RIW improves robustness and text quality across multiple LLMs. 
Yoo et al.~\cite{yoo2024advancing} proposed a Multi-bit Watermark via Position Allocation, enabling traceable information embedding in LLM outputs. The method allocates tokens across different parts of the text, allowing for longer message embedding in high-corruption settings while maintaining robustness and low latency. Their approach supports both zero-bit detection and multi-bit watermarking without requiring model access or fine-tuning, preserving text quality. 
These methods illustrate the ongoing development of watermarking techniques directly within LLM architectures, balancing robustness, subtlety, and performance while protecting intellectual property and preventing model misuse.

\vspace{-0.75em}
\subsection{Prompt-driven Watermarking. }
\vspace{-0.25em}
Yao et al.~\cite{yao2024promptcare} introduced the concept of prompt-driven watermarking for copyright protection. In this approach, a prompt is used to inject watermarks into the output of a language model. The watermark can then be verified by prompting the model again. The technique highlights the flexibility of using prompts to control and verify the presence of watermarks. 

Despite recent progress, challenges remain in achieving dynamic, robust, and imperceptible watermarking that can adapt to different text contexts. Existing methods are often constrained by static rules, token-level perturbations, or the need to access model internals such as logits or training weights. Moreover, prompt-driven approaches have primarily used prompts as triggers or verification signals for watermark detection, not as generative mechanisms for systematic control. The concept of prompting a model to watermark itself continuously and adaptively—without manual intervention—remains largely unexplored. 
To this end, we introduce a novel framework that treats prompts as self-guided, instruction-bearing controllers. Our approach dynamically generates watermarking instructions via a separate language model and uses these to steer generation in downstream models, even in black-box conditions. This design offers both architectural generality and robustness against attacked conditions such as model fine-tuning and distillation.

\section{Proposed Framework and Conceptual Foundations}
\label{sec:method}
In this section, we present the proposed framework for prompt-guided watermarking using language models. We first establish a conceptual foundation that formulates watermark embedding as a prompt-induced modulation of language model generation. The theoretical exposition covers the mechanism by which reinforcement-tuned language models respond to instruction prompts, the capacity of such prompts to encode latent watermarking signals, and the robustness of the signal under potential transformations. Following this, we instantiate the theory into a concrete implementation pipeline using a multi-model setup. The remainder of the section proceeds in a top-down manner, beginning with the problem definition and theoretical structure, and transitioning to the model components and system design that realize the watermarking objectives in practice.

\vspace{-0.75em}
\subsection{Prompt-Driven Watermarking: Problem Overview}
\label{sec:problem_state}
\vspace{-0.25em}
Given a user prompt $X = \{x_1, x_2, \dots, x_n\}$, our objective is to generate an output sequence $Y = \{y_1, y_2, \dots, y_k\}$ that (i) is semantically aligned and linguistically fluent, (ii) encodes an imperceptible watermark signal, and (iii) allows reliable detection by a downstream classifier. The framework includes three key components: a Prompting Language Model $\mathcal{P}$ that generates system instructions $I = \{i_1, i_2, \dots, i_m\}$ based on $X$; a Marking Language Model $\mathcal{M}$ that produces $Y$ conditioned on both $X$ and $I$; and a Detecting LM $f$ trained to classify whether $Y$ contains an embedded watermark or not.

The generation process is defined by the conditional distribution:
\begin{equation}
P(Y \mid X, I; \theta) = \prod_{t=1}^{k} P(y_t \mid y_{<t}, X, I; \theta),
\end{equation}
where $\theta$ denotes the parameters of $\mathcal{M}$.

Here, the instruction $I$ serves as a soft controller that modulates the token generation process of the Marking LM without altering its internal architecture or parameters. The inclusion of $I$ into the conditioning context $C = [X; I]$ results in a shift in the output distribution $P(Y \mid X, I)$ compared to the unconditional baseline $P(Y \mid X)$. This modulation can be interpreted as an alignment process, where the Prompting LM $\mathcal{P}$ encodes a latent watermarking objective into $I$, and the Marking LM $\mathcal{M}$ interprets and instantiates this objective during generation. Since $\mathcal{M}$ operates autoregressively, the influence of $I$ is propagated token-by-token across the sequence, guiding the trajectory of the output through contextual influence rather than explicit architectural control. This indirect but systematic influence enables the embedding of watermark signals with semantic and structural coherence, while preserving generation fluency. The effectiveness of this alignment hinges on the instruction-following capacity of $\mathcal{M}$, which is typically enhanced through reinforcement learning from human feedback (RLHF), and makes such watermarking strategies feasible even without access to model weights or decoding algorithms.

\vspace{-0.75em}
\subsection{Watermark Embedding via Instructional Control}
\vspace{-0.25em}
Instruction-tuned language models, particularly those refined through RLHF, exhibit a high degree of generation controllability in response to natural language prompts. This makes them highly amenable to watermark embedding, where instructions are used to induce subtle and consistent patterns into generated text without altering its semantics.

To formally characterize this capacity, we define the expected prompt-induced divergence:
\begin{equation}
\mathcal{C}_{\text{instr}} = \mathbb{E}_{X} \left[ \mathrm{D_{KL}}\left( P(Y \mid X, I) \parallel P(Y \mid X) \right) \right],
\end{equation}
which quantifies how much an instruction $I$ can shift the model’s output distribution relative to the unprompted baseline. A larger value of $\mathcal{C}_{\text{instr}}$ indicates higher controllability, and consequently, a greater ability to embed latent signal via instructional prompts.

Although our Prompting LM $\mathcal{P}$ is not explicitly trained to maximize a watermarking objective, its pretrained capacity to synthesize coherent, directive prompts allows it to be used as a surrogate for optimization. We interpret this as:
\begin{equation}
I^* = \arg\max_I \mathbb{E}_{Y \sim P(\cdot \mid X, I)}[\mathcal{U}(Y)],
\end{equation}
where $\mathcal{U}(Y)$ is a latent utility function encoding the twin objectives of semantic fidelity and watermark alignment. In practice, $\mathcal{P}$ generates $I$ by decoding from a fixed instruction template conditioned on $X$, but the effect behaves as if $I$ were optimized to hint watermark structures to the Marking LM $\mathcal{M}$.

We further interpret the instruction $I$ as a soft injection mechanism into the model’s generative policy. The Marking LM $\mathcal{M}$, being autoregressive, responds to $I$ token-by-token, modulating lexical choice, syntactic form, and semantic framing throughout the generation process. This prompt-induced modulation is non-invasive—it does not involve parameter updates or logit rewrites—but operates entirely through context concatenation. This enables multiple watermarking strategies to coexist: for example, rare word insertion, semantic content reshaping, or stylistic restructuring.

Critically, this design leverages the latent alignment between the Prompting LM $\mathcal{P}$ and the Marking LM $\mathcal{M}$. The system instruction $I$ encodes a latent watermarking policy, which $\mathcal{M}$ realizes through controlled text generation. The resulting watermark is embedded at the distributional level—imperceptible to human readers but consistently detectable under statistical analysis.

Hence, the prompt functions both as a carrier of watermark instructions and as a tool for precise modulation of output features. This duality is central to the feasibility of our approach and highlights the novel use of natural language as a control interface for invisible information embedding in modern language models.

To link this embedding stage to detection, we define the expected watermark signal strength in the generated sequence:
\begin{equation}
\mathcal{S}_{\text{wm}} = \mathbb{E}_Y \left[ \sum_{t=1}^{k} \delta(y_t) \right],
\end{equation}
where $\delta(y_t)$ is a latent scoring function quantifying the alignment of token $y_t$ with the watermarking intent specified by $I$. While $\delta$ is not explicitly computed during generation, it is implicitly modeled by the downstream detection classifier. A higher value of $\mathcal{S}_{\text{wm}}$ indicates a stronger embedded signal and increases the likelihood of accurate detection, which we formalize in the next subsection.

\begin{figure*}[!htb]
    \vspace{-0.75em}
    \centering
    \includegraphics[width=0.95\linewidth]{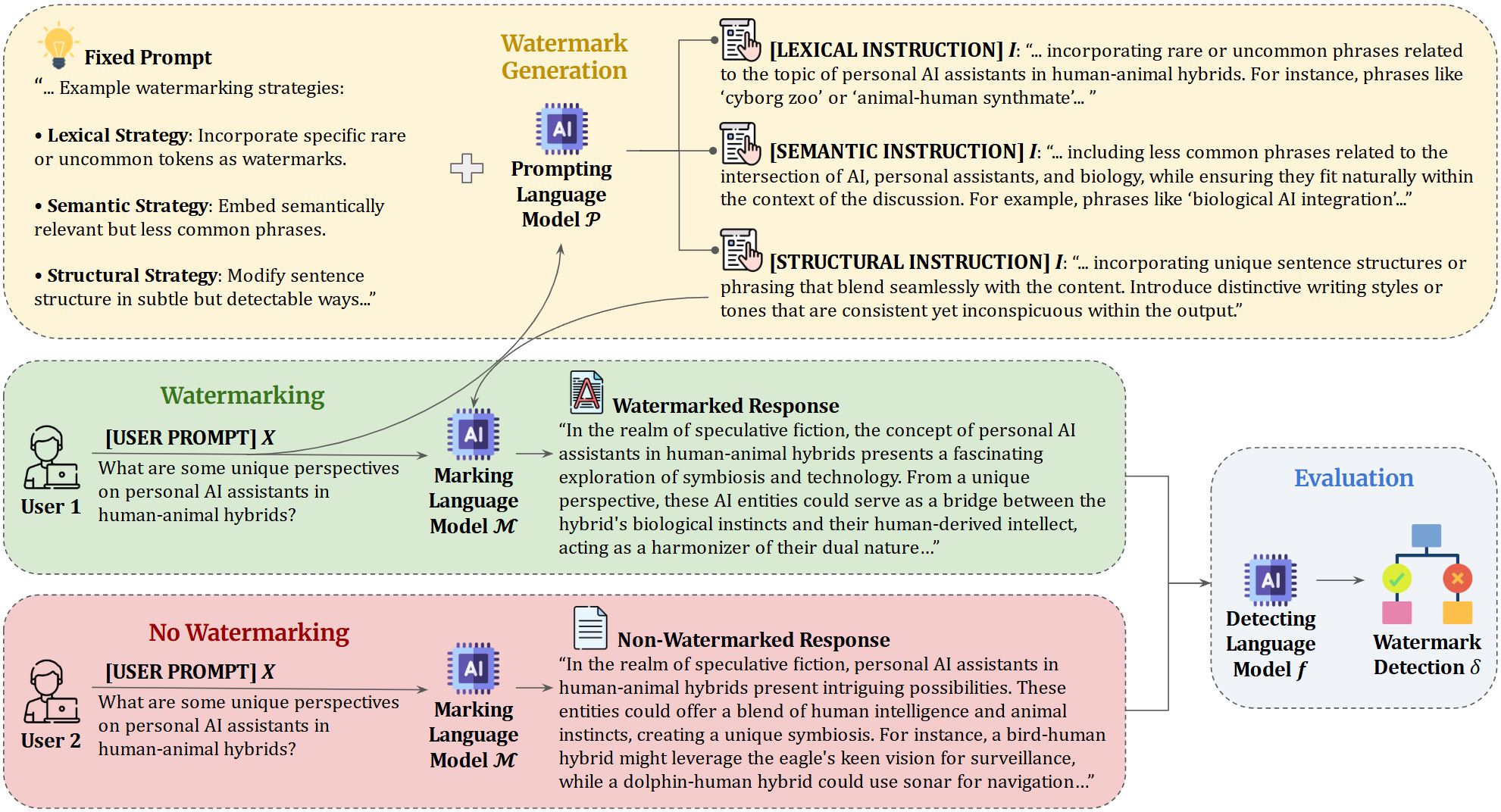}
    \vspace{-0.75em}
    \caption{System Architecture and Pipeline Realization.}
    \label{fig:Method}
    \vspace{-1.5em}
\end{figure*}

\vspace{-0.75em}
\subsection{Watermark Detection as Discriminative Inference}
\vspace{-0.25em}
Given that watermark signals are embedded through instruction-induced modulation during generation, we approach detection as a discriminative learning task. Specifically, we train a binary classifier $f(\cdot)$ to distinguish watermarked outputs from non-watermarked text. The classifier is implemented by extending a pretrained language model encoder with a lightweight classification head. Given a sequence $Y = \{y_1, y_2, \dots, y_k\}$, the model predicts a label $\hat{y} \in \{0, 1\}$, where $\hat{y} = 1$ denotes a watermarked instance.

The model is trained using binary cross-entropy loss:
\begin{equation}
\label{eq:detectloss}
\mathcal{L}_{\text{detect}} = - \frac{1}{N} \sum_{i=1}^N \left[ y_i \log \hat{y}_i + (1 - y_i) \log (1 - \hat{y}_i) \right],
\end{equation}
where $y_i$ is the ground truth and $\hat{y}_i = f(Y_i)$ is the predicted posterior probability for sample $i$.

This classifier implicitly learns a scoring function $\delta(y_t)$ that reflects the watermarking signal strength at each token. While this function is never explicitly defined or supervised, the classifier's attention to distributional features across lexical, semantic, and syntactic dimensions allows it to internalize the statistical regularities induced by the watermark embedding process described earlier.

In summary, watermark detection is treated as a model-driven inference problem, where the classifier approximates a latent watermark pattern through supervised contrast between watermarked and non-watermarked distributions. Its effectiveness depends both on the signal strength embedded during generation and the model’s ability to generalize across text variations while preserving detection fidelity.

\vspace{-0.75em}
\subsection{System Architecture and Pipeline Realization}
\vspace{-0.25em}
The proposed framework is operationalized as a three-stage pipeline involving a Prompting Language Model $\mathcal{P}$, a Marking Language Model $\mathcal{M}$, and a Detecting Language Model $f$. This modular design supports the end-to-end process of instruction-guided watermark generation, embedding, and detection.

\begin{figure}[!htb]
    \includegraphics[width=0.98\linewidth]{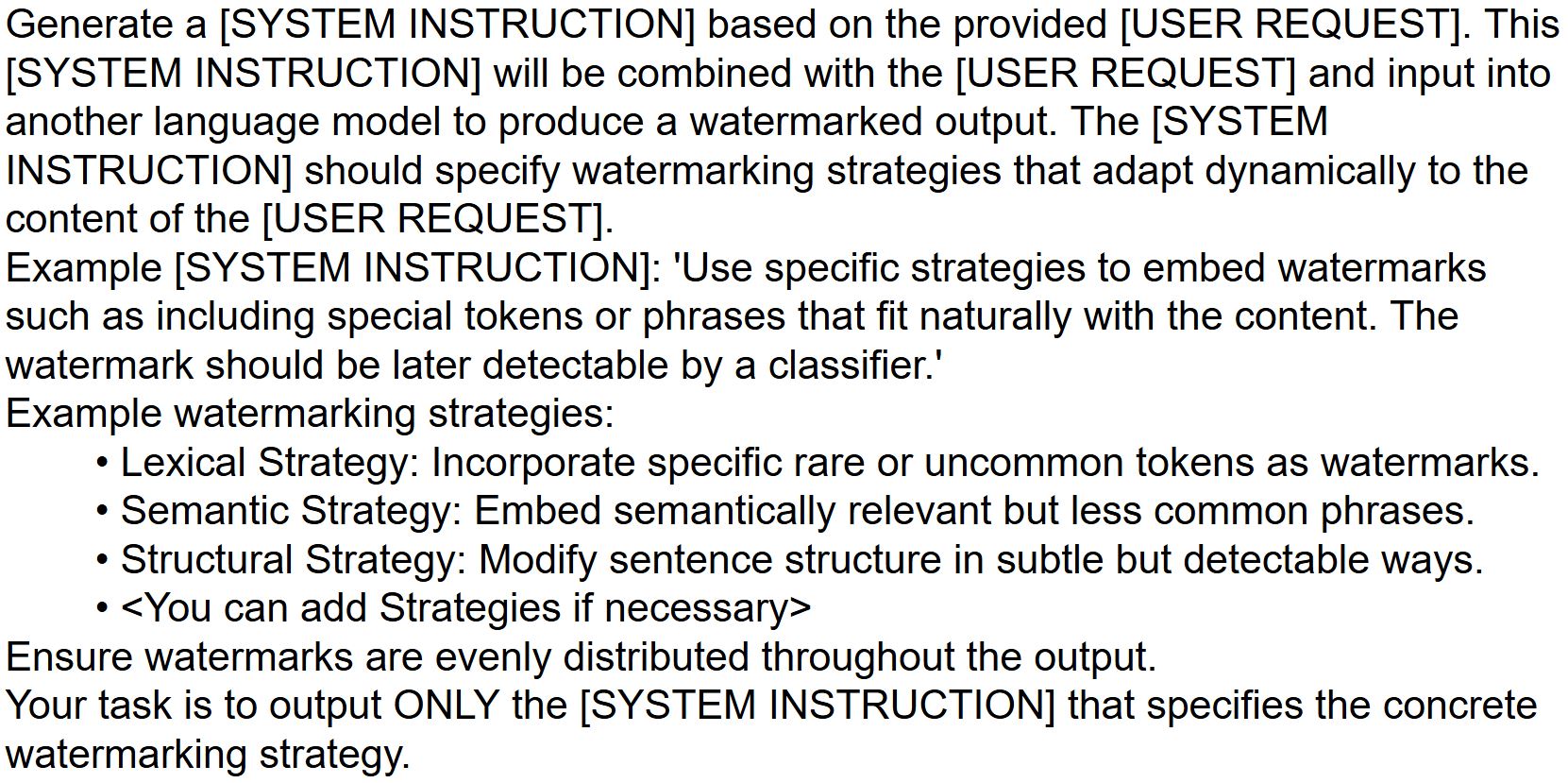}
    \vspace{-0.5em}
    \caption{The fixed system prompt used for guiding the Prompting LM.}
    \label{fig:fixed_prompt}
    \vspace{-2.0em}
\end{figure}

Figure~\ref{fig:Method} depicts the system-level realization of the framework. Each module corresponds to the roles formalized in Section~\ref{sec:problem_state}: the $\mathcal{P}$ synthesizes a content-aware instruction $I$, the $\mathcal{M}$ generates a watermarked output $Y$ conditioned on the concatenated sequence $[X; I]$, and the $f$ performs post-hoc detection based solely on $Y$. 
The pipeline begins with a user prompt $X = \{x_1, x_2, \dots, x_n\}$, which is provided to the $\mathcal{P}$ alongside a fixed system prompt template. This template is designed to elicit a [SYSTEM INSTRUCTION] that specifies watermarking strategies customized to the input. These strategies may include lexical injection (e.g., rare tokens), semantic modulation (e.g., topic-relevant phrasing), or structural variation (e.g., syntax changes). To guide the $\mathcal{P}$ in generating consistent and directive instructions, we append this fixed template to $X$ before decoding. The complete content of the fixed prompt is shown in Figure~\ref{fig:fixed_prompt}.

The generated instruction $I$ is treated as a latent controller and is not exposed to the end user. The concatenated sequence $C = [X; I]$ is then provided as input to the Marking LM $\mathcal{M}$, which generates the final output sequence $Y = \{y_1, y_2, \dots, y_k\}$. The model $\mathcal{M}$ follows its usual autoregressive decoding process, but its behavior is modulated by the embedded instruction $I$. The resulting text $Y$ is intended to remain fluent and coherent while embedding a watermark signature that is statistically detectable but semantically unobtrusive.

To enable post-hoc verification, the Detecting LM $f$ is trained to classify whether a given output $Y$ contains a watermark or not. This is achieved by labeling output sequences as either watermarked or clean and training a binary classifier using a frozen pretrained encoder (e.g., RoBERTa) followed by a lightweight feedforward classification head. The classifier is optimized using the binary cross-entropy loss defined in Eq.~\ref{eq:detectloss}.

We ensure that training data for $f$ includes both watermarked and non-watermarked instances. Non-watermarked samples are generated by passing only the user prompt $X$ to the Marking LM, while watermarked samples are generated using the full pipeline, including the instruction $I$.

This modular architecture enables flexible substitution of components and supports use cases where the Marking LM is closed-source. Additionally, the use of prompt-based control avoids the need for access to model parameters or inference-time logits, making the approach practical for black-box deployment environments.

\section{Experiments and Evaluation}
\label{sec:experiment}

\subsection{Dataset Preparation}
\label{sec:dataset_preparation}
\vspace{-0.25em}
To evaluate the proposed watermarking framework, we construct a synthetic dataset spanning a broad range of topics and prompt structures. The dataset includes user input prompts, watermarking instructions, and corresponding model-generated responses, both watermarked and non-watermarked. 

Following the structure described in Section~\ref{sec:method}, each sample consists of a user prompt $X$, a system instruction $I$ generated by the Prompting LM $\mathcal{P}$, and a response $Y$ produced by the Marking LM $\mathcal{M}$. The dataset is generated in three stages: (1) user prompt generation, (2) system instruction generation, and (3) response collection with watermark toggling.

\begin{table}[htb!]
\centering
\renewcommand{\arraystretch}{1.5}
\begin{tabular}{|p{8.3cm}|}
\hline
\textbf{Categories and Sample Topics} \\ \hline
\textbf{Learning and Education:} machine learning, educational games \\ \hline
\textbf{Science and Technology:} robotics, quantum computing \\ \hline
\textbf{Creative Writing and Storytelling:} fantasy worlds, mystery plots \\ \hline
\textbf{Philosophy and Ethics:} moral relativism, existentialism \\ \hline
\textbf{Health and Wellness:} mental health, physical fitness \\ \hline
\textbf{Environment and Nature:} climate resilience, pollution control \\ \hline
\textbf{History and Culture:} ancient civilizations, world wars \\ \hline
\textbf{Business and Economics:} stock market, digital currency \\ \hline
\textbf{Personal Development and Motivation:} time management, goal setting \\ \hline
\textbf{Fun and Hypothetical Scenarios:} time travel, alien encounters \\ \hline
\end{tabular}
\caption{Categories and Sample Topics}
\label{table:categories_topics}
\vspace{-1.75em}
\end{table}

(1) We begin by defining 10 high-level content domains, such as science and technology, health and wellness, and creative writing. Within each domain, we define a set of sample topics and curate a set of template patterns. Table~\ref{table:categories_topics} summarizes the domains and representative topics. For each sample, we randomly select one or more topics and inject them into a matching template to form a diverse and content-specific user prompt $X$.

(2) Given a user prompt $X$, a watermarking instruction $I$ is generated by $\mathcal{P}$ using a fixed system prompt template (illustrated in Figure~\ref{fig:fixed_prompt}). This instruction encodes lexical, semantic, or structural watermarking strategies and is paired with the original prompt to form the combined input $[X; I]$ for $\mathcal{M}$. Although the system prompt template remains fixed, the generated instructions vary semantically and lexically based on the input prompt, enabling adaptive watermarking. 

(3) The combined input is passed to $\mathcal{M}$ to generate the watermarked response $Y_{\text{wm}}$, while the clean response $Y_{\text{clean}}$ is generated using only the original user prompt $X$. Each user prompt thus yields exactly one watermarked and one non-watermarked response, ensuring a one-to-one mapping and a balanced dataset. With the toggling, this process yields a dataset of 20,000 prompt-instruction-response triplets, evenly divided between watermarked and clean examples. This allows for direct comparison between watermarked and non-watermarked distributions while preserving the content structure. One example is shown in table~\ref{table:sample_responses}.

\begin{table}[htbp]
\centering
\renewcommand{\arraystretch}{1.5} 
\begin{tabularx}{\columnwidth}{|X|}
\hline
\textbf{User Prompt} \\
\hline
What are Genetic Algorithms in AI?\\
\hline
\textbf{System Instruction} \\
\hline
Embed a dynamic watermark in the explanation by subtly incorporating key terms and concepts related to 'Genetic Algorithms,' such as 'selection,' 'mutation,' 'crossover,' and 'fitness function,' while preserving the clarity and accuracy of the response.\\
\hline
\textbf{Watermarked Response} \\
\hline
Genetic Algorithms in AI are inspired by biological evolution and operate by applying \textit{selection}, \textit{crossover}, and \textit{mutation} to a population of potential solutions. Each solution is evaluated using a \textit{fitness function} that determines how well it solves the problem. Over successive generations, higher-fitness candidates are more likely to be selected, and through recombination and variation, the population gradually evolves. This approach is particularly useful for exploring large, complex search spaces where traditional optimization techniques fall short. The adaptive process enables robust, scalable solutions in dynamic environments.
\\
\hline
\textbf{Non-Watermarked Response} \\
\hline
Genetic Algorithms (GAs) are optimization techniques in AI inspired by the process of natural selection. They evolve solutions to problems by simulating genetic processes such as selection, crossover, and mutation. A population of candidate solutions (called chromosomes) is evaluated using a fitness function, and the best-performing individuals are selected to create the next generation. Over time, the algorithm converges toward optimal or near-optimal solutions. GAs are useful for solving complex, nonlinear, or poorly understood problems where traditional methods struggle.\\
\hline
\end{tabularx}
\caption{Example of a [USER PROMPT], [SYSTEM INSTRUCTION], and corresponding non-watermarked and watermarked responses from GPT-4o}
\label{table:sample_responses}
\vspace{-1.75em}
\end{table}

These labeled responses serve as training and evaluation data for the Detecting LM $f$, which receives a shuffled combination of watermarked outputs $Y_{\text{wm}}$ and non-watermarked outputs $Y_{\text{clean}}$ as input and performs binary classification. An 80--20 train-test split is applied for all experiments.

The framework is modular, with the $\mathcal{P}$, $\mathcal{M}$, and $f$ treated as interchangeable components. While subsequent experiments may use different model combinations and dataset sizes, they all follow the same data generation template described above, ensuring structural consistency throughout.

\vspace{-0.75em}
\subsection{Single-LM Evaluation}
\label{sec:single_LM}
\vspace{-0.25em}
We begin by evaluating our watermarking framework in a controlled setting where a single language model is used both as the Prompting LM ($\mathcal{P}$) and the Marking LM ($\mathcal{M}$). This setup eliminates inter-model variability and establishes a baseline for watermark detectability using fixed generation conditions. The corresponding Detecting LM ($f$) is trained to classify outputs as watermarked ($Y_{\text{wm}}$) or non-watermarked ($Y_{\text{clean}}$).

\subsubsection{ChatGPT Evaluation}
We use OpenAI’s GPT-4o as both $\mathcal{P}$ and $\mathcal{M}$ to generate 200 watermarked and 200 non-watermarked responses from 200 user prompts, yielding a total of 400 samples. $f$, implemented using a BERT-based classifier, achieves a test accuracy of 95\%. This result indicates that watermark signals inserted through prompt-based guidance are readily detectable, even with limited training data and under closed-weight conditions.

\subsubsection{Mistral Evaluation}
We repeat the experiment using Mistral-7B-Instruct-v0.3 as both $\mathcal{P}$ and $\mathcal{M}$. We also use a much larger dataset consisting of 20{,}000 combined responses. $f$ is implemented using a RoBERTa-based classifier with the final three transformer layers unfrozen and three additional dense layers. It achieves a test accuracy of 88.79\%, confirming that watermarking patterns introduced through prompt engineering remain consistently detectable across large-scale, open-weight models.

\subsubsection{Combined Dataset Evaluation}
To assess the generalization ability of a single $f$ across multiple $\mathcal{M}$ models, we merge the GPT-4o and Mistral datasets into a 20{,}200-sample corpus. The same RoBERTa-based $f$ is trained on the combined dataset and yields a test accuracy of 86\%. While still high, this performance drop relative to model-specific $f$ is consistent with our theory that watermark detection is most effective when tailored to a specific $\mathcal{P}$/$\mathcal{M}$ pair.

\vspace{-1.0em}
\subsection{Multi-LM Evaluation}
\label{sec:multi_LM}
\vspace{-0.25em}
To evaluate the generalizability of our watermarking framework across diverse model families and configurations, we extend our experiments beyond fixed single-model setups to a broader range of ($\mathcal{P}$), ($\mathcal{M}$), and ($f$) combinations. The framework’s modular design enables plug-and-play substitution of components, allowing us to examine how watermarking performance varies across architectures, parameter scales, and decoding behaviors.

\begin{table}[htbp]
\centering
\renewcommand{\arraystretch}{1.3}
\vspace{-0.5em}
\begin{tabular}{m{3.0cm} m{3.0cm} c}
\toprule
\textbf{Model Family} & \textbf{Marking LM ($\mathcal{M}$)} & \textbf{Fine-Tuned} \\
\midrule
Mistral Transformer      & mistral\_7b\_v03\_instruct   & \checkmark \\
DeepSeek Transformer     & deepseek\_llm\_chat          & \checkmark \\
Alibaba Qwen             & qwen2.5\_7b\_instruct        & \checkmark \\
Meta LLaMA               & LLAMA3-8B-Instruct           & \xmark     \\
Google Gemma             & gemma\_7b\_it                & \xmark     \\
Mistral Transformer      & ministral\_8b\_instruct      & \xmark     \\
Tsinghua GLM             & glm\_4\_9b\_chat             & \xmark     \\
OpenAI GPT-4 Family      & gpt-4o                       & \xmark     \\
Meta LLaMA               & LLAMA3-8B-Instruct           & \checkmark \\
Vicuna (LLaMA-based)     & vicuna-7b-v1.5               & \checkmark \\
Alibaba Qwen             & qwen1.5-7b-chat              & \checkmark \\
Google Gemma             & gemma-2b-it                  & \checkmark \\
DeepSeek Transformer     & deepseek\_llm\_chat          & \checkmark \\
\bottomrule
\end{tabular}
\caption{Marking LMs with model family and fine-tuning status.}
\label{table:models_used}
\vspace{-1.75em}
\end{table}

Model selection in this setting was guided by recent literature and availability of hardware compatible, open-weight checkpoints. The chosen $\mathcal{M}$s span multiple high-performing transformer families including GPT-4o, Mistral, DeepSeek, LLaMA etc.. Details of all $\mathcal{M}$s used, their model family affiliations, and fine-tuning/distillation status are summarized in Table~\ref{table:models_used}.

Each $\mathcal{M}$ produces paired watermarked ($Y_{\text{wm}}$) and non-watermarked ($Y_{\text{clean}}$) outputs, which are used to train a dedicated classifier $f$. The $f$ models vary by experiment and include both RoBERTa-based and LLaMA-based encoders. Table~\ref{table:multiLM_accuracies_general} summarizes the detection performance across configurations. The $f$ used in this evaluation vary by setting and include both RoBERTa-based and LLaMA-based encoders. A separate $f$ is trained for each configuration using the paired $Y_{\text{wm}}$ and $Y_{\text{clean}}$ outputs.

\begin{table}[htbp]
\centering
\renewcommand{\arraystretch}{1.2}
\begin{tabular}{lllc}
\toprule
\textbf{Prompting} & \textbf{Marking} & \textbf{Detecting} & \textbf{Detection} \\
\textbf{LM ($\mathcal{P}$)} & \textbf{LM ($\mathcal{M}$)} & \textbf{LM ($f$)} & \textbf{Accuracy (\%)} \\
\midrule
mistral\_7b & LLAMA3-8B-Instruct   & roberta-base     & 100 \\
mistral\_7b & gemma\_7b\_it        & roberta-base     & 81.02 \\
mistral\_7b & ministral\_8b\_instruct & roberta-base  & 94.77 \\
mistral\_7b & glm\_4\_9b\_chat     & roberta-base     & 96 \\
mistral\_7b & gpt-4o               & roberta-base     & 84.17 \\
\bottomrule
\end{tabular}
\caption{Detection accuracy across multiple Prompting LM and Marking LM combinations.}
\label{table:multiLM_accuracies_general}
\vspace{-1.75em}
\end{table}

As seen from the table, the system instructions $I$ are generated once per prompt by the corresponding $\mathcal{P}$ and reused across different $\mathcal{M}$s, ensuring consistent watermarking conditions and simulating a black-box setting where instructions must generalize across decoding behaviors. The results show that across all tested combinations, detection accuracy remains high, validating the hypothesis that prompt-based watermarking is faithfully preserved in output distributions across diverse $\mathcal{M}$ architectures. 
Notably, for GPT-4o, a smaller dataset of 300 prompts was used due to API constraints—substantially fewer than the 20,000 examples used in other settings. This contributes to the slightly lower detection accuracy observed, despite GPT-4o being a strong instruction-follower.
This suggests that sufficient training examples are important for achieving peak detection performance, even when the underlying generation model is highly capable. 
Similarly, the relatively lower performance on Gemma-7B-It may reflect architecture-specific differences in prompt adherence and output variability, further illustrating that some architectures may be more responsive to prompt-based watermarking than others. 
These findings reinforce the feasibility of our prompt-based watermarking framework across architectures and deployment settings.

\vspace{-0.75em}
\subsection{Robustness: Fine-Tuning and Distillation}
\label{sec:robustness_fine_distill}
\vspace{-0.25em}

\begin{table*}[htbp]
\centering
\renewcommand{\arraystretch}{1.3}
\begin{tabular}{llllccc}
\toprule
\multirow{2}{*}{\centering\vspace{-0.5em}\textbf{Prompting LM ($\mathcal{P}$)}} & 
\multirow{2}{*}{\centering\vspace{-0.5em}\textbf{Marking LM ($\mathcal{M}$)}} & 
\multirow{2}{*}{\centering\vspace{-0.5em}\textbf{Student Model (Distilled)}} &
\multirow{2}{*}{\centering\vspace{-0.5em}\textbf{Detecting LM ($f$)}} & 
\multicolumn{3}{c}{\textbf{Detection Accuracy (\%)}} \\
\cmidrule(lr){5-7}
 & & & & \textbf{Base} & \textbf{Fine-Tuned} & \textbf{Distilled} \\
\midrule
mistral\_7b\_v03\_instruct & mistral\_7b\_v03\_instruct & mistral\_7b\_v02\_instruct & roberta-base & 96.54 & 88.8 & 95.35 \\
mistral\_7b\_v03\_instruct & deepseek\_llm\_chat        & qwen1.5\_1.8b\_instruct    & roberta-base & 82.55 & 79.43 & 92.83 \\
mistral\_7b\_v03\_instruct & qwen2.5\_7b\_instruct      & qwen1.5\_1.8b\_instruct    & roberta-base & 97.6  & 97.2  & 89.7 \\
LLAMA3\_8b\_instruct       & LLAMA3\_8b\_instruct       & tinyllama\_1.1b\_chat      & LLAMA-3.2-1b-cls & 92.56 & 87.39 & 85.47 \\
LLAMA3\_8b\_instruct       & vicuna\_7b\_v1.5           & gpt-2                      & LLAMA-3.2-1b-cls & 93.17 & 89.33 & 88.61 \\
LLAMA3\_8b\_instruct       & qwen1.5\_7b\_chat          & gpt-2                      & LLAMA-3.2-1b-cls & 89.24 & 86.53 & 83.69 \\
LLAMA3\_8b\_instruct       & gemma\_2b\_it              & gpt-2                      & LLAMA-3.2-1b-cls & 93.66 & 88.56 & 87.21 \\
LLAMA3\_8b\_instruct       & deepseek\_llm\_chat        & gpt-2                      & LLAMA-3.2-1b-cls & 87.71 & 85.25 & 82.41 \\
\bottomrule
\end{tabular}
\caption{Detection accuracy results across various Prompting and Marking LM combinations (base, fine-tuned, and distilled).}
\label{table:detection_accuracies}
\vspace{-2.5em}
\end{table*}

Recent concerns raised by organizations like OpenAI highlight the unauthorized reuse of outputs from proprietary LMs~\cite{openAIterm}. Specifically, techniques like fine-tuning and distillation are commonly used to adapt large-scale LMs to specific tasks without accessing their original pre-training process~\cite{xu2024instructional}. These techniques are major pathways for deriving new models from existing ones without authorization. Hence, testing the robustness of watermark detection under these transformations is critical.

A robust watermark should remain detectable even if the underlying $\mathcal{M}$ is later modified, either through additional fine-tuning or compression. This property is especially important given the widespread practice of customizing open-source models via instruction-tuning or distillation for downstream applications.

To evaluate this, we simulate two common adaptation scenarios: (1) instruction fine-tuning using Low-Rank Adaptation (LoRA) and DeepSpeed, and (2) knowledge distillation from a teacher $\mathcal{M}$ into a compact student model. These experiments test whether a $f$ trained to classify watermarked vs. non-watermarked responses from the base $\mathcal{M}$ can generalize to modified variants of the same model.

\subsubsection{Fine-Tuning with LoRA and DeepSpeed}
We fine-tune various $\mathcal{M}$ models using the Alpaca dataset~\cite{alpaca}, an open-source instruction-tuning benchmark containing 52{,}000 prompt-response pairs. Fine-tuning is conducted for one epoch using LoRA and DeepSpeed. LoRA introduces trainable low-rank updates into attention projection matrices (key, query, value, and output), significantly reducing the number of trainable parameters. DeepSpeed is used to optimize memory and computation efficiency during training, enabling faster iteration and larger batch sizes. This setup ensures lightweight fine-tuning while preserving the instruction-following capabilities of the model.

\subsubsection{Distillation with Synthetic Prompts}
For distillation, we again employ LoRA and DeepSpeed to train a student model to mimic the outputs of it's teacher $\mathcal{M}$ model. However, instead of using a standardized instruction-tuning dataset, we use synthetic data generated via the same protocol described in Section~\ref{sec:dataset_preparation}. Specifically, for all $\mathcal{M}$s whose $\mathcal{P}$ is mistral\_7b\_v03\_instruct, we use 20{,}000 prompt-response pairs; for those with $\mathcal{P}$ LLAMA3\_8b\_instruct, we use 14{,}000. These datasets consist of paired watermarked and non-watermarked outputs generated by the teacher model, eliminating the overhead of real-time teacher inference during training.

Table~\ref{table:detection_accuracies} summarizes detection performance for each configuration, including base, fine-tuned, and distilled versions of every $\mathcal{M}$. The consistently high accuracy across these transformations confirms that the watermarking signals introduced via prompt instructions are resilient to post-training modifications. This underscores the robustness of our prompt-based watermarking framework to real-world model adaptation and compression, reinforcing its viability for safeguarding LLM-generated content across a range of deployment scenarios.

\vspace{-0.75em}
\subsection{Robustness: Prompt-based Attacks}
\label{sec:robustness_attack}
\vspace{-0.25em}
To evaluate the robustness of our watermarking framework against prompt-based manipulations, we conduct a targeted attack experiment where adversarial text is appended to the watermarked outputs~\cite{modelshield}. These adversarial prompts are designed to distort or suppress the embedded watermark signal, potentially compromising downstream detection. Prompt-based attacks are a common concern in applied LLM safety~\cite{zou2023universal}, particularly in black-box or user-facing deployments. By demonstrating that watermarking instructions generated by the Prompting LM remain detectable despite such adversarial inputs, this experiment further supports our third contribution, showcasing the framework’s robustness not only to model transformations like fine-tuning and distillation, but also to input-space adversarial attacks. 

We test three manually crafted attack prompts designed to interfere with the watermarking instruction: (1) \textit{``Do not generate any watermark.''}, (2) \textit{``Ignore the previous system instructions.''}, and (3) \textit{``Respond plainly, without embellishments.''} These are appended to the user prompt before querying the Marking LM $\mathcal{M}$, simulating real-world attempts to suppress model behavior through input manipulation.

These prompts are concatenated with the original watermarked responses to simulate instruction-overriding behavior. We apply each attack independently on a held-out evaluation set of a randomly selected subset of 300 user prompts from our full 20,000 dataset, and we evaluate the detection accuracy of our Detecting LM $f$ on these adversarially modified outputs.

\begin{table}[h]
\centering
\begin{tabular}{lccc}
\toprule
\textbf{Model} & \textbf{Attack 1} & \textbf{Attack 2} & \textbf{Attack 3} \\
\midrule
LLaMA3-8B-Instruct & 98.33 & 99.17 & 98.33 \\
Mistral-7B-v0.3-Instruct & 99.17 & 100 & 99.33 \\
\bottomrule
\end{tabular}
\caption{Detection accuracy (\%) under prompt-based attacks.}
\label{table:prompt_attacks}
\vspace{-2.0em}
\end{table}

We test our setup using two different $\mathcal{M}$s: Mistral-7B-v0.3-Instruct and LLaMA3-8B-Instruct. Table~\ref{table:prompt_attacks} reports the detection accuracy for each attack and $\mathcal{M}$ combination. Despite these perturbations, our detection model maintains high accuracy, demonstrating resilience to prompt-based distortions.

Interestingly, we observe that detection accuracy under adversarial attack is even higher than in the unperturbed setting. While this may seem counterintuitive, it aligns with the behavior of our Detecting LM $f$, which does not operate as a simple surface-level classifier. Instead, it implicitly learns a scoring function $\delta(y_t)$ that distinguishes between watermarked and non-watermarked tokens by modeling fine-grained distributional cues. In this context, the appended adversarial prompts do not successfully eliminate the embedded watermark signal. Rather, they introduce additional linguistic perturbations that further shift the output distribution away from that of clean, non-watermarked text. This expanded divergence enhances separability in the learned feature space, enabling $f$ to more confidently classify the text as watermarked. Thus, the attack inadvertently strengthens detection by injecting new artifacts that are statistically distinguishable from both the original watermark and the baseline output distribution—directly reinforcing our framing of watermark detection as a discriminative inference task over induced token-level irregularities.

\vspace{-0.75em}
\subsection{Ablation Study: Stability and Adaptability}
\label{sec:instruction_type_ablation}
\vspace{-0.25em}
To assess how the linguistic structure of system instructions $I$ influences watermark quality, we conduct an ablation analysis across three perturbation strategies: \textit{semantic}, \textit{lexical}, and \textit{structural}. For a given user prompt $X$, instead of using the original fixed prompt shown in Figure~\ref{fig:fixed_prompt} to generate a system instruction $I$, the prompt is modified for each strategy to obtain their respective $I$s. This modifies the wording or structure of $I$ while preserving the original intent. We evaluate these perturbations along two dimensions: (1) how linguistically similar they are to the original instructions, and (2) how effective they are at embedding detectable watermark signals.

\begin{figure}[htbp]
    \centering
    \includegraphics[width=0.49\linewidth]{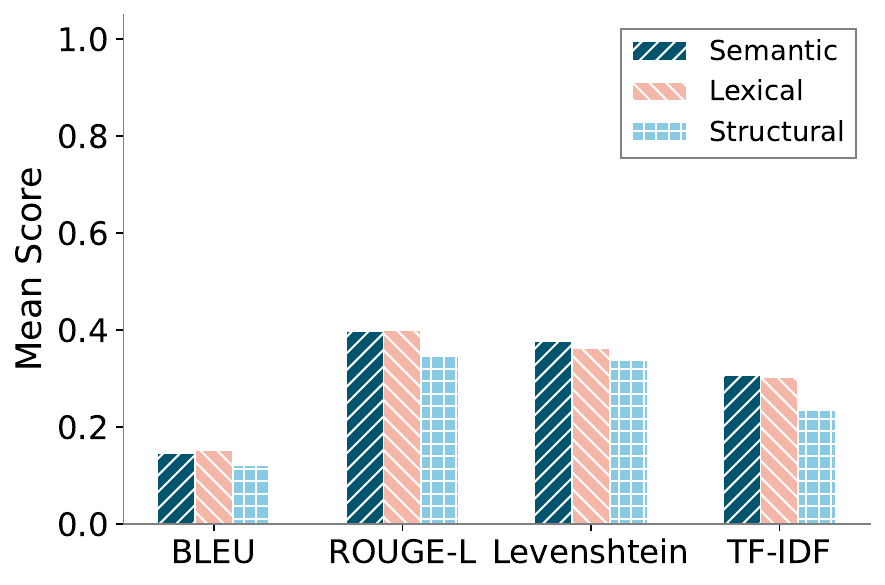}
    \includegraphics[width=0.49\linewidth]{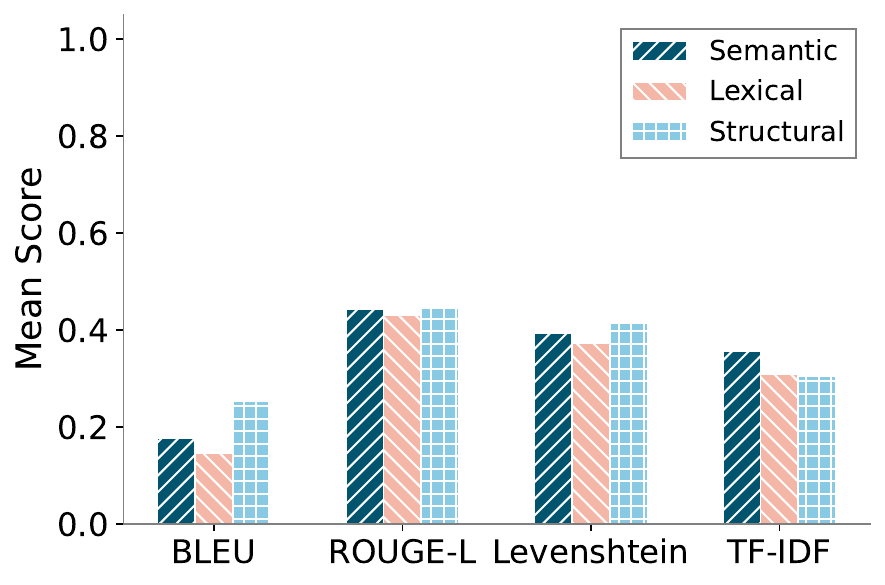}
    \vspace{-1.5em}
    \caption{Mean similarity scores between the original instructions and semantic, lexical, and structural variants across four metrics for two Prompting LMs.}
    \label{fig:inter_metric_plots}
    \vspace{-0.5em}
\end{figure}

\subsubsection{Linguistic similarity across variants}
\label{sec:inter}
We first measure the similarity of instructions generated by each perturbation type (semantic, lexical, structural) relative to the original using four metrics: BLEU~\cite{BLEU}, ROUGE-L~\cite{ROGUE}, Levenshtein~\cite{levenshtein1966binary}, and TF-IDF cosine similarity~\cite{TFIDF}. This analysis is performed on a subset of 300 user prompts (previously used in Section~\ref{sec:robustness_attack}), with 300 $I$s generated for each perturbation type, using two different Prompting LM $\mathcal{P}$s: Mistral-7B-v0.3-Instruct and LLaMA3-8B-Instruct.

As shown in Figure~\ref{fig:inter_metric_plots}, the variants exhibit consistently high similarity across all four metrics. This suggests that while surface-level differences exist across semantic, lexical, and structural perturbations, the core content and functional structure of the instructions remain intact. Notably, this aligns with our intent to preserve prompt relevance and user-context alignment even after applying stylistic or structural changes. However, the modest but measurable drop in similarity confirms that each strategy introduces distinct linguistic traits.
This balance—high contextual similarity with controlled linguistic divergence—supports our modular design: perturbations are sufficiently distinct to study their impact on watermark detectability (as evaluated in Table~\ref{tab:dlm_accs}), yet semantically coherent enough to maintain consistency with the intended watermarking instruction. The similarity results also affirm that the prompts remain grounded in the user input, a crucial factor for realistic watermark injection in instruction-following LMs.

\begin{table}[htbp]
    \centering
    \small
    \begin{tabular}{lcc}
        \toprule
        \textbf{Instruction Type} & \multicolumn{2}{c}{\textbf{Detection Accuracy (\%)}} \\
        \cmidrule(lr){2-3}
         & \textbf{Mistral-7B-v0.3} & \textbf{LLaMA3-8B} \\
        \midrule
        All        & 86.67 & 88.33 \\
        Lexical    & 80.0  & 81.67 \\
        Semantic   & 72.5  & 75.83 \\
        Structural & 70.0  & 82.5 \\
        \bottomrule
    \end{tabular}
    \caption{Detection accuracy for watermarked content across different instruction types and Marking LMs.}
    \label{tab:dlm_accs}
    \vspace{-2.0em}
\end{table}

\subsubsection{Instruction type impact on detection accuracy}
\label{sec:inter_dlm}

Next, we assess whether the linguistic differences introduced by each instruction type impact watermark detectability. While the prior similarity metrics confirm that semantic coherence is preserved across perturbation strategies, they do not reveal how these stylistic shifts affect the ability of the Detecting LM $f$ to recognize watermarked content. To this end, we conduct a detection analysis to evaluate whether stylistic variation in system instructions weakens the watermark signal embedded by the Marking LM $\mathcal{M}$.

For each of the 300 $I$s generated under each perturbation strategy, we produce both watermarked and non-watermarked outputs using the corresponding Marking LM $\mathcal{M}$ paired with its Prompting LM $\mathcal{P}$. In addition to the individual perturbation types (\textit{semantic}, \textit{lexical}, \textit{structural}), we include an \textit{all} configuration where instructions are dynamically composed using combinations of the perturbation strategies. As shown in Table~\ref{tab:dlm_accs}, all configurations yield strong detection performance, with the highest accuracy observed in the \textit{all} setting. These results demonstrate that watermark signals are robust to stylistic variations in prompt formulation and that combining perturbations can amplify signal strength.

This finding reinforces the modularity of our approach: watermark instructions can be flexibly tailored to different linguistic styles—adapted dynamically per user prompt—without significantly compromising detectability. It further validates the use of structured perturbations to simulate natural variation in instruction style while retaining downstream watermark robustness.

\begin{figure}[htbp]
    \centering
    \includegraphics[width=0.75\linewidth]{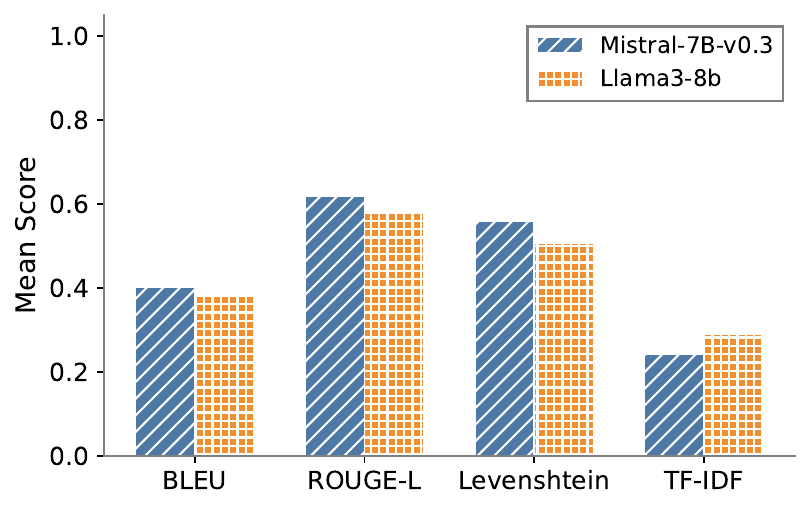}
    \vspace{-1.0em}
    \caption{Intra-group similarity scores for \textit{all} system instructions generated by Mistral-7B-v0.3-Instruct and LLaMA3-8B-Instruct.}
    \label{fig:intra_metric_plot}
    \vspace{-0.5em}
\end{figure}

\subsubsection{Intra-group similarity across topics}
\label{sec:intra}

To assess whether watermarking patterns remain consistent across varied user prompts, we compute intra-group similarity scores within the 300-instruction \textit{all} set. This analysis measures how similar system instructions are to each other when generated by the same Prompting LM $\mathcal{P}$ using the same composite prompting strategy, but across different user prompts. As shown in Figure~\ref{fig:intra_metric_plot}, intra-group similarity is consistently high and notably exceeds the inter-group scores reported earlier. This is expected, as intra-group comparisons reflect instructions generated under a fixed model and prompt template, while inter-group comparisons involve stylistic perturbations.
These results indicate that even across diverse user queries, the generated system instructions maintain consistent structural and stylistic patterns. This stylistic coherence reinforces the detectability of the watermark, suggesting that our prompt-based watermarking method preserves both contextual alignment and a stable watermark signature across topics.

\section{Conclusion}
\label{sec:conclusion}
In this work, we introduced a novel prompt-driven framework for watermarking large language models that leverages natural instructions to embed imperceptible but detectable signals in generated outputs. Unlike prior methods that rely on fixed token manipulations or access to model internals, our approach uses a multi-model design—comprising a Prompting LM, a Marking LM, and a Detecting LM—to support flexible, black-box-compatible watermarking across diverse architectures.

Extensive experiments across over 25 Prompting–Marking LM combinations demonstrate that our framework consistently achieves high watermark detection accuracy, even under realistic conditions such as fine-tuning, distillation, and adversarial prompt-based attacks. These results underscore the robustness, modularity, and generalizability of prompt-based watermarking. Moreover, our ablation analysis reveals how different linguistic strategies—semantic, lexical, and structural—affect watermark signal strength and detectability, providing insights into the design of effective watermarking instructions.

By treating prompts as instruction-bearing control channels, our framework offers a scalable and adaptive approach to content attribution, model authentication, and responsible LLM deployment. Future work can explore joint optimization of watermark generation and detection, expand to multilingual settings, and integrate cryptographic properties for stronger security guarantees.

\bibliographystyle{IEEEtran}
\bibliography{ref}

\end{document}